\documentclass[10pt,twocolumn,letterpaper]{article}

\usepackage{wacv}
\usepackage{times}
\usepackage{epsfig}
\usepackage{graphicx}
\usepackage{amsmath}
\usepackage{amssymb}
\usepackage{amssymb}
\usepackage{multirow}
\usepackage{mathtools}
\usepackage{subcaption}

\DeclarePairedDelimiter\floor{\lfloor}{\rfloor}

\wacvfinalcopy 

\def\wacvPaperID{463} 

\ifwacvfinal\fi
\begin{document}

\title{Weakly Supervised Temporal Action Localization Using Deep Metric Learning}

\author{Ashraful Islam \\
 Rensselaer Polytechnic Institute\\
{\tt\small islama6@rpi.edu}
\and
Richard J.~Radke \\
Rensselaer Polytechnic Institute\\
{\tt\small rjradke@ecse.rpi.edu}
}

\maketitle
\ifwacvfinal\thispagestyle{empty}\fi

\begin{abstract}
Temporal action localization is an important step towards video understanding. Most current action localization methods depend on untrimmed videos with full temporal annotations of action instances. However, it is expensive and time-consuming to annotate both action labels and temporal boundaries of videos. To this end, we propose a weakly supervised temporal action localization method that only requires video-level action instances as supervision during training. We propose a classification module to generate action labels for each segment in the video, and a deep metric learning module to learn the similarity between different action instances. We jointly optimize a balanced binary cross-entropy loss and a metric loss using a standard backpropagation algorithm. Extensive experiments demonstrate the effectiveness of both of these components in temporal localization. We evaluate our algorithm on two challenging untrimmed video datasets: THUMOS14 and ActivityNet1.2. Our approach improves the current state-of-the-art result for THUMOS14 by 6.5\% mAP at IoU threshold 0.5, and achieves competitive performance for ActivityNet1.2. 
\end{abstract}

\section{Introduction} \label{sec:intro}

Video action recognition and action localization are active areas of research. There are already impressive results in the literature for classifying action categories in trimmed videos \cite{i3d, Varol2018LongTermTC, Tran2018ACL}, and important contributions have been made in action localization in untrimmed videos \cite{ssn, Wang2016TemporalSN, frcnn}. Temporal action localization is a much harder task than action recognition due to the lack of properly labelled datasets for this task and the ambiguity of temporal extents of actions \cite{Schindler2008ActionSH}. Most current temporal action localization methods are fully supervised, i.e., the temporal boundaries of action instances must be known during training. However, it is very challenging to create large-scale video datasets with such temporal annotations. On the other hand, it is much easier to label video datasets with only action instances, since billions of internet videos already have some kind of weak labels attached. Hence, it is important to develop algorithms that can localize actions in videos with minimum supervision, i.e., only using video-level labels or other weak tags. 

\begin{figure*}[t]
\begin{center}
\includegraphics[width=\textwidth]{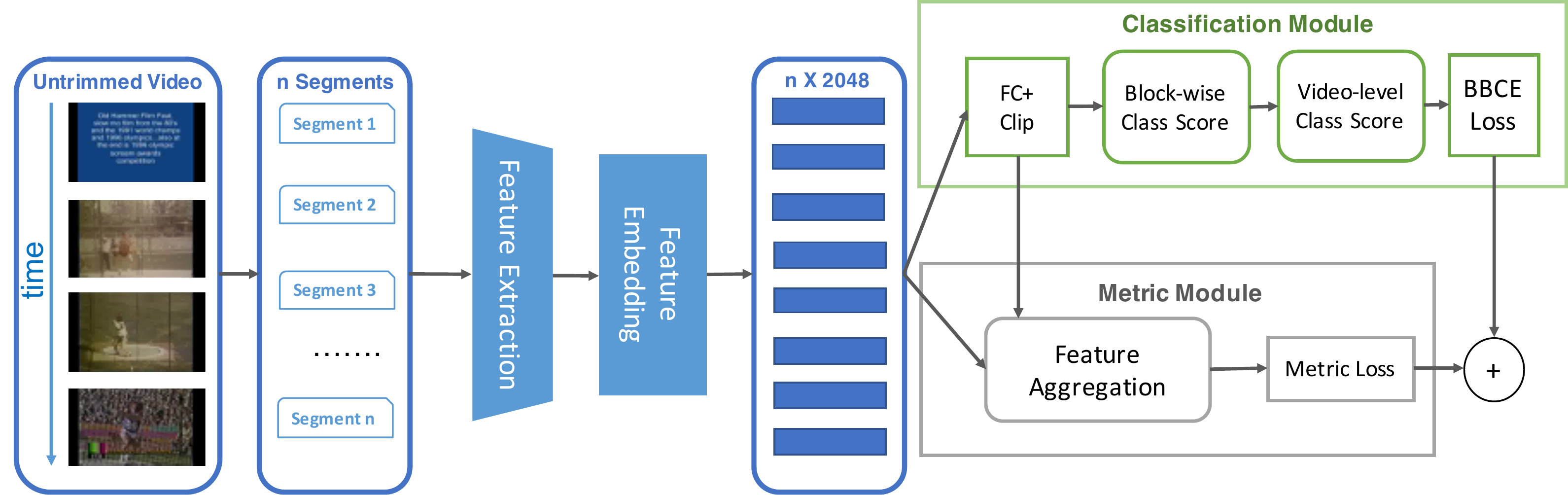}
\end{center}
    \caption{Our algorithm extracts features from video segments and feeds them into classification and metric learning modules. We optimize these jointly to learn the network weights.}
    \label{fig:model}
\end{figure*}

In this paper, we propose a novel deep learning approach to temporally localize actions in videos in a weakly-supervised manner. Only the video-level action instances are available during training, and our task is to learn a model that can both classify and localize action categories given an untrimmed video. To achieve this goal, we propose a novel classification module and a metric learning module. Specifically, given an untrimmed video, we first extract equal-length segments from the video, and obtain segment-level features by passing them through a feature extraction module. We feed these features into a classification module that measures segment-level class scores. To calculate the classification score of the whole video, we divide the video into several equal-length blocks, combine the block-level classification scores to get the video-level score, and then apply a balanced binary cross-entropy loss to learn the parameters. To facilitate the learning, we also incorporate a metric learning module. We propose a novel metric function to make frames containing the same action instance closer in the metric space, and frames containing different classes to be farther apart. We jointly optimize the parameters of both of these modules using the Adam optimizer \cite{Kingma2015AdamAM}. An overview of our model is shown in Fig.~\ref{fig:model}.

The proposed method exhibits outstanding performance on the  THUMOS14 dataset \cite{THUMOS14}, outperforming the current state of the art by 6.5\% mAP at IoU threshold 0.5, and showing comparable results even to some fully-supervised methods. Our method also achieves competitive results on the ActivityNet1.2 \cite{activitynet} dataset.

\section{Related Work}\label{sec:related_works}

\textbf{Video Action Analysis.}
There has been significant progress in the field of action recognition and detection, particularly due to the introduction of large-scale datasets \cite{THUMOS14, activitynet,charades, hmdb, AVA, Soomro2012UCF101AD} and the development of deep learning models. For example, two-stream networks \cite{simonyan2014two}, 3D convolutional networks (C3D) \cite{c3d} and recently I3D networks \cite{i3d} have been extensively applied to learn video representations and have achieved convincing performance. For temporal action localization, various deep learning based methods include temporal segment networks \cite{Wang2016TemporalSN}, structured segment networks \cite{ssn}, predictive-corrective networks \cite{Dave2017PredictiveCorrectiveNF}, and TAL-Net \cite{frcnn}. Most of these techniques use temporal annotations during training, while we aim to use only video-level labels for action localization.   

\textbf{Deep Metric Learning.}
The objective of metric learning is to learn a good distance metric such that the distance between the same type of data is reduced and the distance between different types of data is enlarged. Traditional metric learning approaches rely on linear mapping to learn the distance metric, which may not capture non-linear manifolds in complex tasks like face recognition, activity recognition, and image classification. To solve this problem, kernel tricks are usually adopted \cite{Xiong2014PersonRU, Lu2013ImageSC}. However, these methods cannot explicitly obtain nonlinear mappings, and also suffer from scalability problems. With the advent of deep learning, deep neural network-based approaches have been used to learn non-linear mappings in metric learning. For example, Hu \etal~\cite{Hu2014DiscriminativeDM} trained a deep neural network to learn hierarchical non-linear mappings for face verification. Bell and Bala \cite{Bell2015LearningVS} learned visual similarity using contrastive embedding \cite{Hadsell2006DimensionalityRB}. Schroff \etal~\cite{facenet} used triplet embedding \cite{tripletintro} on faces for face verification and clustering.

\textbf{Weakly-Supervised Temporal Localization.}
Weakly supervised deep learning methods have been widely studied in object detection \cite{Bilen2016WeaklySD, Cinbis2017WeaklySO, Oquab2015IsOL}, semantic segmentation \cite{Hong2017WeaklySS, Khoreva2017SimpleDI}, visual tracking \cite{Zhong2010VisualTV}, and video summarization \cite{Hong2017WeaklySS}. However, there are only a few weakly supervised methods in temporal action localization that rely only on video-level labels during training. It should be noted that there are different types of weak supervision for the temporal localization task. For example, some works use movie scripts or subtitles as weak supervision \cite{Bojanowski2013FindingAA, Duchenne2009AutomaticAO}, whereas others use the temporal order of actions during training \cite{Richard2017WeaklySA, Huang2016ConnectionistTM}. We do not use any information about temporal ordering in our model. Our approach only uses a set of action classes for each video during training.

Wang \etal~\cite{wang2017untrimmednets} proposed a model named UntrimmedNets consisting of a classification module that predicts the classification scores for each video clip and a selection module that detects important video segments. The algorithm uses a Softmax function to generate action proposals, which is not ideal for distinguishing multiple action classes. It is also based on a temporal segments network \cite{Wang2016TemporalSN} that considers a fixed number of video segments, which is not effective for variable-length video datasets. Nguyen \etal~\cite{stpn} added a sparsity-based loss function and class-specific action proposals (contrary to class-agnostic proposals in UntrimmedNets). However, the sparsity constraint for attention weights that they propose would hurt localization performance in videos that contain very few background activities.  

Shou \etal~\cite{shou2018autoloc} introduced Outer-Inner-Contrastive Loss to automatically predict the temporal boundaries of each action instance. Paul \etal~\cite{wtalc} proposed techniques that combine Multiple Instance Learning Loss with Co-activity Similarity Loss to learn the network weights. Our proposed method is similar to this work with novel contributions in several important areas.  In particular, we adopt a block-based processing strategy to obtain a video-level classification score, and propose a novel metric function as a similarity measure between activity portions of the videos.
Su \etal~\cite{su2018weakly} proposed shot-based sampling instead of uniform sampling and designed a multi-stage temporal pooling network for action localization. Zeng \etal~\cite{zeng2019breaking} proposed an iterative training strategy to use not only the most discriminative action instances but also the less discriminative ones. Liu \etal~\cite{liu2019completeness} recently proposed a multi-branch architecture to model the completeness of actions, where each branch is enforced to discover distinctive action parts. They also used temporal attention similar to \cite{stpn} to learn the importance of video segments, showing a minor performance improvement over \cite{wtalc}.

\section{Proposed Algorithm} \label{sec:proposed_algo}
In this section, we introduce the detailed pipeline of our proposed algorithm. We first describe the data processing and feature extraction modules. We then present the classification and deep metric learning modules and introduce loss functions to jointly optimize them \footnote{Code accompanying this paper is available at https://github.com/asrafulashiq/wsad.git}. 

\textbf{Problem Formulation.}
We consider an untrimmed video as a collection of segments, where each segment contains an equal number of frames. Let a video $\mathbf{V}$ be represented as a collection of segments $\{c_i\}_{i=1}^n$, where $n$ is the total segment length, and an associated activity class set with $n_c$ unique activity instances represented as $\mathbf{a} = \{a^k\}_{k=1}^{n_c}$, where $a^k \in \mathcal{A}$, the set of all action classes in the dataset. 
The training data set contains $N$ videos $\{\mathbf{V}_i\}_{i=1}^{N}$ with their associated labels $\{\mathbf{a}_i\}_{i=1}^{N}$. The length and activity instances in the video can vary significantly, and we only have video-level labels during the training period. 
Given a test video, the model will predict a set of action labels with corresponding start time, end time and confidence score. 

\subsection{Feature Extraction}
We extract segment-level features $\{\mathbf{x}_i\}_{i=1}^{n}$, where $\mathbf{x}_i \in \mathbb{R}^{d}$ is a $d$-dimensional feature vector, and $n$ is the segment length of the video. Two-stream networks have become common for action recognition and detection \cite{i3d, feichtenhofer2016convolutional}. Following \cite{stpn}, we use the I3D network \cite{i3d} pretrained on the Kinetics dataset \cite{kinetics} to extract features from each video segment. Both the RGB and optical flow streams are used for feature extraction, and we fuse them together to get a single feature vector for each video segment. We use the TV-L1 algorithm \cite{TVL1} to extract the flow. We do not use any fine-tuning on this feature extractor network.

\subsection{Feature Embedding}
Given a feature representation of a video $\mathbf{V}$ as $\{\mathbf{x}_i\}_{i=1}^{n}$, we feed the features to a module consisting of a fully connected layer followed by a ReLU and a dropout layer. This module modifies the original features extracted from the pre-trained feature extraction module into task-specific embedded features. We keep the dimension of the embedded features the same as the dimension of the extracted features. The embedded features are denoted by $\{ \mathbf{u}_{i}\}_{i=1}^{n}$, where $\mathbf{u}_i \in \mathbb{R}^{d}$.  

\subsection{Classification Module} \label{subsec:class}
Next, we learn a linear mapping $\mathbf{W}_f \in \mathbb{R}^{C\times d}$ and bias $\mathbf{b} \in \mathbb{R}^{C}$ followed by a clipping function $\varphi_{\kappa}(\cdot)$ to obtain class-specific activations $s_{i} \in \mathbb{R}^{C}$ for each segment, where $C$ is the total number of class labels, i.e.,
\begin{align} \label{eq:cam}
    s_i = \varphi_\kappa(\mathbf{W}_f\mathbf{x_i}+\mathbf{b})
\end{align}
where $\varphi_\kappa(\cdot)$ is defined by 
\begin{equation*}
    \varphi_\kappa(x) = \begin{cases}
        \kappa & \text{if } x > \kappa\\
        -\kappa & \text{if } x < -\kappa\\
        x & \text{otherwise}
    \end{cases}
\end{equation*}
The necessity of using a clipping function is discussed in Sec.~\ref{sec:metric_learning}. \\

To obtain the video-level classification score, we use a block-based processing strategy. Specifically, since the total segment length $n$ of a video $\mathbf{V}$ can vary, we divide the video into blocks, where  each block is a set of an equal number of consecutive segments, i.e., $\mathbf{V}=\{B_i\}_{i=1}^{n_{B}(\mathbf{V})}$, where $n_{B}(\mathbf{V})=\floor*{\frac{n}{l_w}}$ is the total number of blocks, and $l_w$ is the number of segments in each block. We empirically chose the value of $l_w$ (discussed in Sec.~\ref{sec:ablation}).

We calculate $P(c\mid\mathbf{V})$, the probability of the video $\mathbf{V}$ containing particular class $c$, as 

\begin{align} \label{eq:class-score}
    P(c\mid\mathbf{V}) &= P\left(c \mid \{B_i\}_{i=1}^{n_{B}(\mathbf{V})}\right) \\ 
    &= 1 - \prod_{i=1}^{n_B({\mathbf{V}})} (1 - P(c \mid B_i))
\end{align}

\noindent where $P(c \mid B_i)$ is the probability that the $i$-th block contains class $c$. One approach to obtain this probability is to pick the highest class activation in that block. However, an activity would likely cover several video segments. Hence, following \cite{wtalc}, we compute the average of the $k$-max class activation scores in the block as
\begin{equation}
    P(c\mid B_i) =  \sigma\left(\frac{1}{k} \max_{\substack{
    l \subset \mathcal{I}_i
    }} \sum_{j=1}^k  s_{l_j}^c\right)
\end{equation}
where $\mathcal{I}_i$ contains the segment indices for the $i$-th block, $\sigma(\cdot)$ is the sigmoid activation function, and $ s_{l_j}^c$ is the class activation score for the $l_j$-th segment.

We compute $P(c\mid \mathbf{V})$ for each class $c \in \{1, 2, \ldots, C\}$. As a video can contain multiple activities, this is a multi-label classification problem. Hence, the binary cross-entropy loss (BCE) is an obvious choice. However, we found through experiments that the standard BCE loss performs poorly in this case, mainly due to the class-imbalance problem. Xie and Tu \cite{Xie2015HolisticallyNestedED} first introduced a class-balancing weight to offset the class-imbalance problem in binary cross entropy. Similar to them, we introduce a balanced binary cross-entropy loss, which produces better results in practice. We calculate the balanced binary cross-entropy (BBCE) loss as
\begin{align}\label{eq:BBCE}
    \mathcal{L}_{\text{BBCE}} =\ & \frac{\sum_{c=1}^{C}y_c \log P(c \mid \mathbf{V})}{\sum_{c=1}^{C}y_c} \\
    + & \frac{\sum_{c=1}^{C}(1-y_c) \log(1- P(c \mid \mathbf{V}))}{\sum_{c=1}^{C}(1-y_c)}
\end{align}

Here, $y_c$ is set to $1$ if the video contains class $c$, otherwise it is set to $0$. The effectiveness of $\mathcal{L}_{\text{BBCE}}$ is demonstrated in Sec.~\ref{sec:experiments}. 

\subsection{Metric Learning Module} \label{sec:metric_learning}
Here, we first give a brief review of distance metric learning, and how it is incorporated in our algorithm.

\textbf{Distance Metric Learning.} The goal of metric learning is to learn a feature embedding to measure the similarity between input pairs. Let $\mathbf{X}=\{\mathbf{x}_i\}_{i=1}^n$ be input features and $\mathbf{Y}=\{y_i \}_{i=1}^n$ be corresponding labels.  We want to learn a distance function
$D(\mathbf{x}_i, \mathbf{x}_j) = f(\theta; \mathbf{x}_i, \mathbf{x}_j )$, where $f$ is the metric function and $\theta$ is a learnable parameter. Various loss functions have been proposed to learn this metric. Contrastive loss \cite{Chopra2005LearningAS, Hadsell2006DimensionalityRB} aims to minimize the distance between similar pairs and penalize the negative pairs that have distance less than margin $\alpha$:
\begin{align}\label{eq:contrastive_loss}
    \mathcal{L}_{\text{contrastive}}(\mathbf{x}_i, \mathbf{x}_j) = \ & \mathbf{1}(y_i=y_j)D^2(\mathbf{x}_i, \mathbf{x}_j) \ +\\  & \mathbf{1}(y_i \ne y_j ) [\alpha - D^2(\mathbf{x}_i, \mathbf{x}_j)]_{+}
\end{align}
where $[\cdot]_{+}$ indicates the hinge function $\max (0, \cdot)$. 

On the other hand, triplet loss \cite{tripletintro} aims to make the distance of a negative pair larger than the distance of a corresponding positive pair by a certain margin $\alpha$. Let $\{ \mathbf{x}_i^a, \mathbf{x}_i^{p}, \mathbf{x}_i^{n}\}$ be a triplet pair such that $\mathbf{x}_i^a$ and $\mathbf{x}_i^{p}$ have the same label and $\mathbf{x}_i^a$ and $\mathbf{x}_i^{n}$ have different labels. The triplet loss is defined as:
\begin{equation} \label{eq:base_metric_loss}
\mathcal{L}_{\text{triplet}}(\mathbf{x}_i^a, \mathbf{x}_i^p, \mathbf{x}_i^n) =[ D^2(\mathbf{x}_i^a, \mathbf{x}_i^{p}) - D^2(\mathbf{x}_i^a, \mathbf{x}_i^{n}) + \alpha ]_{+}
\end{equation}

\textbf{Motivation. } A set of videos that have similar activity instances should have similar feature representations in the portions of the videos where that activity occurs. On the other hand, portions of videos that have different activity instances should have different feature representations. We incorporate the metric learning module to apply this characteristic in our model.

\textbf{Our Approach.} We use embedded features and class-activation scores to calculate the aggregated feature for a particular class. Let $\mathcal{B}_c = \{ \mathbf{V}_k \}_{k=1}^{N}$ be a batch of videos containing a common class $c$. After feeding the video segments to our model, we extract embedded features $\{ \mathbf{u}_{k,i} \}_{i=1}^{n_k}$ and class activation scores $\{s_{k,i}\}_{i=1}^{n_k}$ for the $k$-th video, where $n_k$ is the length of the video. Following \cite{wtalc}, we calculate the aggregated feature vector for class $c$ from video $\mathbf{V}_k$ as follows:
\begin{align*}
    \mathbf{z}_k^c = \sum_{i=1}^{n_k} \pi_{k,i}^c \mathbf{u}_{k,i} \quad\text{and}\quad
    \mathbf{z}_k^{\neg c} = \sum_{i=1}^{n_k} \frac{1-\pi_{k,i}^c}{n_k-1} \mathbf{u}_{k,i}
\end{align*}
where $\pi_{k,i}^c = \frac{\exp(s_{k,i}^c)}{\sum_{i^\prime=1}^{n_k} \exp(s_{k,i^\prime}^c)}$. Here, $s_{k,i}^c$ is the class activation of the $i$-th segment for class $c$ in video $\mathbf{V}_k$. Hence, $\mathbf{z}_k^c$ is aggregated from feature vectors that have high probability of containing class $c$, and $\mathbf{z}_k^{\neg c}$ is aggregated from feature vectors that have low probability of containing class $c$.
We normalize these aggregated features to a $d$-dimensional hypersphere to calculate $\mathbf{\Tilde{z}}_k^c$ and $\mathbf{\Tilde{z}}_k^{\neg c}$, i.e. $|| \mathbf{\Tilde{z}}_k^c ||_2=1$ and $||\mathbf{\Tilde{z}}_k^{\neg c}||_2=1 $. Here, we can see the motivation behind applying a clipping function in Eqn.~\ref{eq:cam}. If the clipping function is not applied, there might be a segment $i_h$ with a very high class score $s_{k, i_h}^c$, and the value of $\pi_{k, i_h}^c$, which is the output of a Softmax function, will be close to 1 for that segment and  close to 0 for other segments. Hence, the aggregated features will be calculated mostly from the segment with maximum class score, even though there are other segments that can have high class score for a particular class. Therefore, we apply a clipping function to limit the class score to have a certain maximum and minimum value. 

Next, the average distances for positive and negative pairs from a batch of videos with common class $c$ are calculated as
\begin{align*}
    d^{+,c} &= \frac{1}{n_k(n_k-1)} \sum_{\substack{1\le j,j^\prime \le n_k \\ j \ne j^\prime}} D_c^2(\mathbf{\Tilde{z}}_j^c, \mathbf{\Tilde{z}}_{j^\prime}^{ c}), \quad \\
    d^{-,c} &= \frac{1}{n_k(n_k-1)} \sum_{\substack{1\le j,j^\prime \le n_k \\ j \ne j^\prime}} D_c^2(\mathbf{\Tilde{z}}_j^c, \mathbf{\Tilde{z}}_{j^\prime}^{\neg c}) 
\end{align*}

Instead of using cosine distance as the distance function, our intuition is that $D_c$ should be different for different classes, and hence we define 
$
D_c(\mathbf{u}, \mathbf{v}) = || \mathbf{W}_f^c (\mathbf{u} - \mathbf{v})  ||_2
$, 
where $\mathbf{W}_f^c \in \mathbb{R}^{1 \times d}$ is the $c$-th row of the weight matrix of the final fully-connected layer of our model.
To clarify why this is a proper distance function in this case, we can write $D_c(\cdot, \cdot)$ as:
\begin{align} \label{eq:metric2}
    D_c(\mathbf{u}, \mathbf{v}) &= \sqrt{
    (\mathbf{u}-\mathbf{v})^{^\top} (\mathbf{W}_f^c)^{^\top}\mathbf{W}_f^c (\mathbf{u}-\mathbf{v}) 
    } \\
    &= \sqrt{
    (\mathbf{u}-\mathbf{v})^{^\top} \mathbf{M}^c (\mathbf{u}-\mathbf{v}) 
    } \label{eq:metric_general}
\end{align}
where $\mathbf{M}^c = (\mathbf{W}_f^c)^{^\top}\mathbf{W}_f^c$ is a symmetric positive semi-definite matrix. Hence, Eqn.~\ref{eq:metric2} is actually a Mahalanobis type distance function, where the metric $\mathbf{M}^c$ is calculated from the weights of a neural network. Additionally, the class score for class $c$ is calculated from the weight $\mathbf{W}_f^c$; hence $\mathbf{M}^c$ is a metric that can be used in the distance measure only for class $c$. We show in the ablation studies that our proposed distance function is a better metric in this setting.

Finally, we calculate  either the triplet loss $\mathcal{L}_{\text{triplet}}^c = [d^{+, c}-d^{-, c} + \alpha]_{+}$
or contrastive loss $\mathcal{L}_{\text{contrastive}}^c = d^{+, c} + [\alpha - d^{-,c}]_+$ as the metric loss function. We found through experiments that triplet loss performs slightly better than contrastive loss. Hence, we use triplet loss unless stated otherwise. 

\subsection{Temporal Localization}
Given an input test video, we obtain the segment level class score $y_i^{c} = \sigma(s_i^{c})$ where $\sigma(\cdot)$ is the sigmoid function, and calculate the video-level class score $\bar{y}^c$ for each class $c$ following Eqn.~\ref{eq:class-score}. For temporal localization, we detect action instances for each class in a video separately. Given class scores $y_i^c$  for the $i$-th segment and class $c$, we first discard all segments that have class score less than threshold 0.5. The one-dimensional connected components of the remaining segments denote the action instances of the video. Specifically, each action instance is represented by $(i_s, i_e, c, q)$ where $i_s$ is the start index, $i_e$ is the end index, $c$ is the action class, and $q$ is the class score calculated as $q = \max ( \{y_i^c \}_{i=i_s}^{i_e} )+ \gamma  \bar{y}^c$, where $\gamma$ is set to 0.7.

\section{Experiments} \label{sec:experiments}
In this section, we first describe the benchmark datasets and evaluation setup. Then, we discuss implementation details and comparisons of our results with state-of-the-art methods. Finally, we analyze different components in our algorithm. 

\subsection{Datasets and Evaluation}
We evaluate our method on two popular action localization datasets, namely THUMOS14 \cite{THUMOS14} and ActivityNet1.2 \cite{activitynet}, both of which contain untrimmed videos (i.e., there are many frames in the videos that do not contain any action). 

The THUMOS14 dataset has 101 classes for action recognition and 20 classes for temporal localization. As in the literature \cite{stpn, shou2018autoloc, wtalc}, we use 200 videos in the validation set for training and 213 videos in the testing set for evaluation. Though this dataset is smaller than ActivityNet1.2, it is challenging since some videos are relatively long, and it has on average around 15.5 activity segments per video. The length of activity also varies significantly, ranging from less than a second to minutes.

The ActivityNet1.2 dataset has 100 activity classes consisting of 4,819 videos for training, 2,383 videos for validation, and 2,480 videos for testing (whose labels are withheld). Following \cite{wang2017untrimmednets}, we train our model on the training set and test on the validation set. 

We use the standard evaluation metric based on mean Average Precision (mAP) at different intersection over union (IoU) thresholds for temporal localization. Specifically, given the testing videos, our model outputs a ranked list of localization predictions, each of which consists of an activity category, start time, end time, and confidence score for that activity. If a prediction has correct activity class and significant overlap with a ground truth segment (based on the IoU threshold), then the prediction is considered to be correct; otherwise, it is regarded as a false positive.

\subsection{Implementation Details}
We first sample a maximum of 300 segments of a video, where each segment contains 16 frames with no overlap. If the video contains more than 300 segments, we sample 300 segments from the video randomly. Following \cite{stpn}, we use a two-stream I3D network to extract features from each stream (RGB and flow), and obtain 2048-dimensional feature vectors by concatenating both streams. The total loss function in our model is:
\begin{equation}
    \mathcal{L} = \mathcal{L}_{\text{BBCE}} + \lambda \mathcal{L}_{\text{metric}}
\end{equation}
We set $\lambda=1$. We use $\alpha=3$ in the metric loss function, block size $l_w=60$, and $k=10$ (Section~\ref{subsec:class}). For the videos that have total segment length less than 60, we set $l_w$ to be equal to the total segment length and $k$ to be $\min(10, l_w)$. We use batch size 20 with 4 different activity instances per batch such that at least 5 videos have the same activity. The network is trained using the Adam optimizer \cite{Kingma2015AdamAM} with learning rate $10^{-4}$.

\subsection{Comparisons with State-of-the-Art}\label{sec:results}
We compare our result with state-of-the-art fully-supervised and weakly-supervised action localization methods on the THUMOS14 dataset in Table~\ref{tab:thumos_result}. Our method outperforms other approaches by a significant margin. In particular, it achieves 6.5\% more mAP than the current best result at IoU threshold 0.5, and consistently performs better at other thresholds as well.  Our approach even outperforms several fully-supervised methods, though we are not using any temporal information during training. 

Table~\ref{tab:anet_result} shows our result on the ActivityNet1.2 validation set. Here, we see the performance is comparable with the state-of-the-art. We achieve state-of-the-art performance on IoU 0.1 and 0.3, and the results on other IoUs are very close to the current best results. Due to the significant difference between these two datasets, our algorithm does not produce as impressive results for ActivityNet1.2 as it does for THUMOS14 at all IoU thresholds. However, the THUMOS14 dataset has a large number of activity instances per video (around 15 instances per video) compared to ActivityNet1.2 which has only 1.5 instances per video. Moreover, THUMOS14 contains around 71\% background activity per video (compared to 36\% in ActivityNet1.2). Due to the high concentration of activity instances and large background activity, we think THUMOS14 is a better dataset for evaluating the performance of weakly supervised action detection. Therefore, we will concentrate mostly on THUMOS14 for evaluating our algorithm.

\begin{table}
\fontsize{8}{9.5}\selectfont
    \centering
    \caption{Comparison of our algorithm with other state-of-the-art methods on the THUMOS14 dataset for temporal action localization.}
    \begin{tabular}{|c|c|l l l l|}
    \hline
    \multirow{2}{*}{Supervision} & \multirow{2}{*}{Method} & \multicolumn{4}{c|}{IoU}\\ \cline{3-6}
    && 0.1 & 0.3 & 0.5 & 0.7 \\
    \hline
    
    \multirow{5}{*}{Full} & S-CNN \cite{MSCNN} & 47.7 & 36.3 & 19.0 & 5.3 \\
    & CDC \cite{Shou2017CDCCN} & - & 40.1 & 23.3 & 7.9 \\
    & R-C3D \cite{Xu2017RC3DRC} & 54.5 & 44.8 & 28.9 & - \\
    & CBR-TS \cite{Gao2017CascadedBR} & 60.1 & 50.1 & \bf{31.0} & \bf{9.9}\\
    & SSN \cite{ssn} & \bf{60.3} & \bf{50.6} & 29.1 & - \\
    \hline
    \multirow{10}{*}{Weak} & Hide-and-Seek \cite{Singh2017HideandSeekFA} & 36.4 & 19.5 & 6.8 & -\\
    & UntrimmedNets \cite{wang2017untrimmednets} & 44.4 & 28.2 & 13.7 & -\\
    & STPN \cite{stpn} & 52.0 & 35.5 & 16.9 & 4.3 \\
    & AutoLoc \cite{shou2018autoloc} & - & 35.8 & 21.2 & 5.8 \\
    & W-TALC \cite{wtalc} & 55.2 & 40.1 & 22.8 & 7.6\\
    & Su \etal~\cite{su2018weakly} & 44.8 & 29.1 & 14.0 & - \\
    & Liu \etal~\cite{liu2019completeness} & 57.4 & 41.2 & 23.1 & 7.0\\
    & Zeng \etal~\cite{zeng2019breaking} & 57.6 & 38.9 & 20.5 & - \\
    \cline{2-6}
    & Ours & \bf{62.3} & \bf{46.8} & \bf{29.6} & \bf{9.7} \\ 
    \hline
    \end{tabular}
    \label{tab:thumos_result}
\end{table}

\begin{table}[htbp!]
    \fontsize{8}{9.5}\selectfont
    \centering
    \caption{Comparison of our algorithm with other state-of-the-art methods on the ActivityNet1.2 validation set for temporal action localization.}
    \begin{tabular}{|c|c|l l l l|}
    \hline
    \multirow{2}{*}{Supervision} & \multirow{2}{*}{Method} & \multicolumn{4}{c|}{IoU}\\ \cline{3-6}
    && 0.1 & 0.3 & 0.5 & 0.7 \\
    \hline
     Full & SSN \cite{ssn} & - & - & 41.3 & 30.4 \\
    \hline
    \multirow{4}{*}{Weak} & UntrimmedNets \cite{wang2017untrimmednets} &  - & - & 7.4 & 3.9 \\
    &  AutoLoc \cite{shou2018autoloc} & - & - & 27.3 & \bf{17.5} \\
    & W-TALC \cite{wtalc} & 53.9 & 45.5 & \bf{37.0} & 14.6 \\
    & Liu \etal~\cite{liu2019completeness} & - & - & 36.8 & - \\
    \cline{2-6}
    & Ours & \bf{60.5} & \bf{48.4} & 35.2 & 16.3\\
    \hline
    \end{tabular}
    \label{tab:anet_result}
\end{table}

\subsection{Ablation Study} \label{sec:ablation}
In this section, we present ablation studies of several components of our algorithm. We use different values of hyperparameters that give the best result for each architectural change. We perform all the studies in this section using the THUMOS14 \cite{THUMOS14} dataset.

\begin{figure}[htbp!]
\begin{center}
    \includegraphics[width=0.4\textwidth]{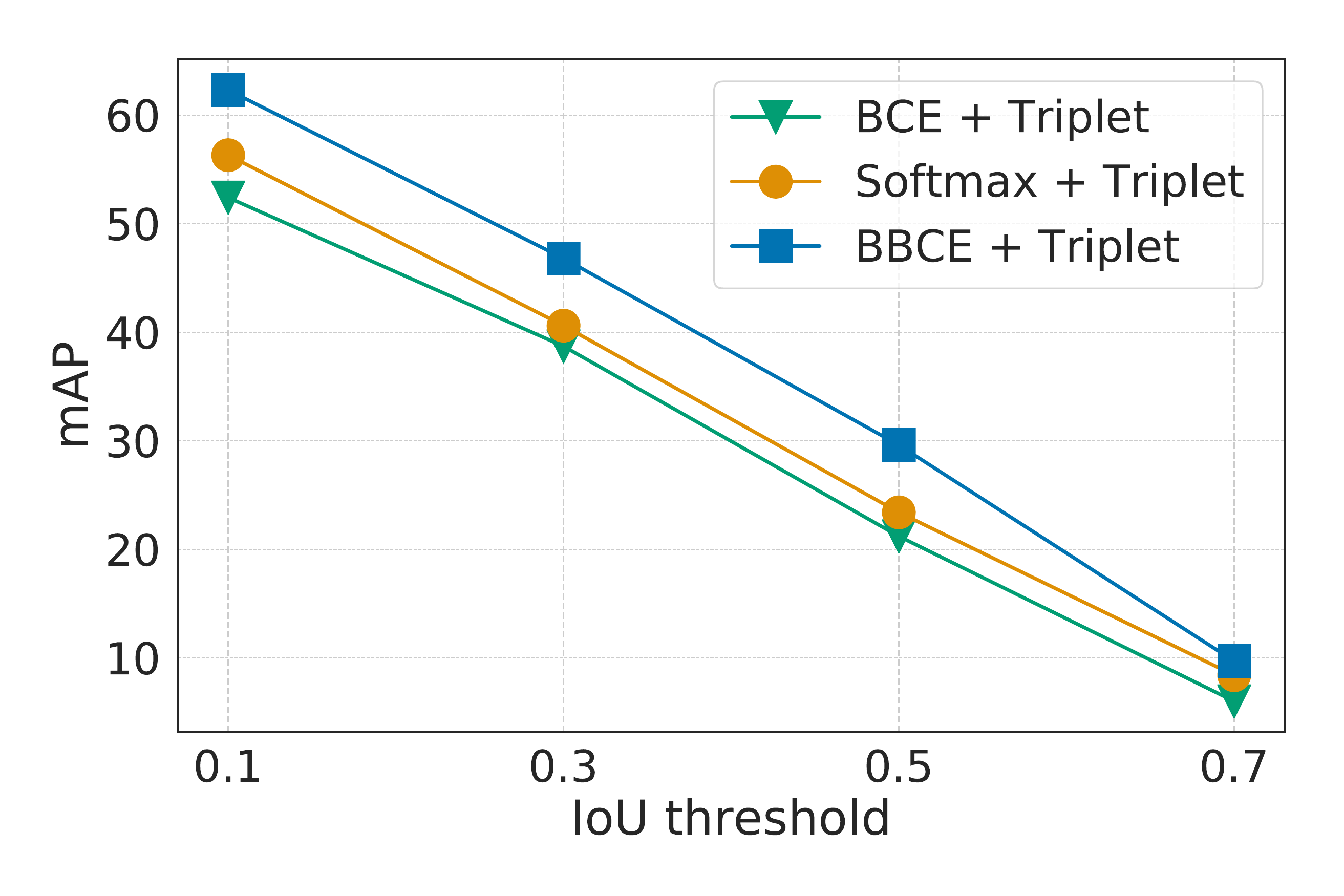}
\end{center}
\vspace{-0.9cm}
    \caption{The mAP performance at different IoU thresholds on the THUMOS14 dataset for different classification loss functions. For the same metric loss function, BBCE performs better than BCE and Softmax loss. Here the Softmax loss is calculated according to the multiple-instance learning loss in \cite{wtalc}.}
    \label{fig:loss_comp}
\end{figure}

\textbf{Choice of classification loss function.} As discussed in Sec.~\ref{subsec:class}, we use the balanced binary cross-entropy (BBCE) loss instead of binary cross-entropy (BCE) and softmax loss. Figure~\ref{fig:loss_comp} presents the effectiveness of BBCE loss over other choices. The same block-based processing strategy for the classification module is also included in the experiment. Our intuition is that the BBCE loss gives equal importance to both foreground activities and background activities, so it can solve the class imbalance problem in a video more accurately.

\textbf{Effect of metric learning module.} To clarify, the goal of using a distance function here is to introduce an extra supervising target, which is especially useful in the weakly-supervised setting. In Table \ref{tab:abl_metric}, we show the performance of our model without any metric loss, with contrastive metric loss, and with triplet loss, respectively. We see significant increases in the overall performance when metric loss is applied. In particular, the average mAP increases by 13.17\% when the contrastive metric loss is applied and 13.32\% when the triplet loss is applied.  

\begin{table}[!htbp]
\fontsize{8.5}{11.5}\selectfont
    \centering
        \caption{Experiments to show the effect of metric function on the THUMOS14 testing set for different IoU thresholds. Here, `Avg' denotes the average mAP over IoU thresholds 0.1, 0.3, 0.5 and 0.7.}
        \vspace{-0.3cm}
    \begin{tabular}{|c|l l l l|l|}
    \hline

      \multirow{2}{*}{Method} & \multicolumn{5}{c|}{IoU}\\ \cline{2-6}
    & 0.1 & 0.3 & 0.5 & 0.7 & Avg  \\
    \hline
    Ours, $\mathcal{L}_{\text{BBCE}}$ & 48.7 & 29.3 & 14.0 & 3.1 & 23.78\\
    Ours, $\mathcal{L}_{\text{BBCE}} + \mathcal{L}_{\text{Contrastive}}$ & 61.7 & 46.6 & 28.4 & 9.3 & 36.95\\
    Ours, $\mathcal{L}_{\text{BBCE}} + \mathcal{L}_{\text{Triplet}}$ & \bf62.3 & \bf46.8 & \bf29.6 & \bf9.7 & \bf37.10\\
    \hline
    \end{tabular}
    \label{tab:abl_metric}
\end{table}

\begin{figure}[htbp!]
\begin{center}
    \includegraphics[width=0.35\textwidth]{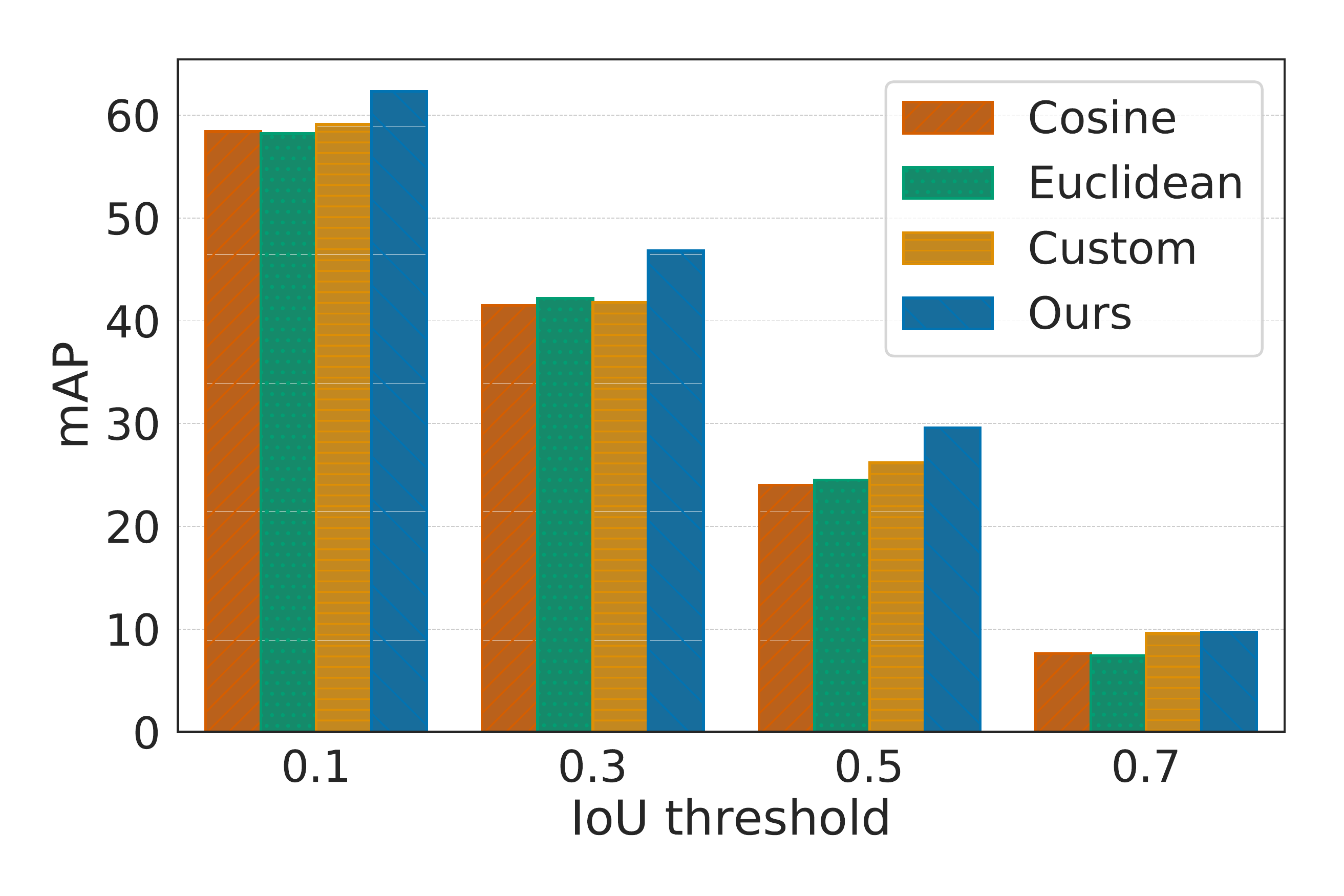}
\end{center}
\vspace{-0.7cm}
    \caption{Performance comparison on the same dataset for different distance functions. Our metric performs better than the cosine distance, Euclidean distance, and a custom learnable distance.}
    \label{fig:distance_comp}
\end{figure}

To validate the effectiveness of our proposed metric over other metric functions, we perform experiments by replacing our distance function with cosine distance, Euclidean distance, and a custom learnable distance function. For the custom distance function, we propose a learnable parameter $M \in \mathbb{R}^{C \times d \times d}$, which is updated through back-propagation, where $C$ is the total number of classes, and set the metric $\mathbf{M}^c = M(c, :, :)$ in Eq.~\ref{eq:metric_general}. Recall that when $\mathbf{M}^c = \mathbf{I}_d$, where $\mathbf{I}_d$ is the $d$-dimensional identity matrix, the metric function becomes the Euclidean distance function. In Fig.~\ref{fig:distance_comp}, we present the results for different distance functions. From the figure, we see that the performances of cosine distance and Euclidean distance are quite similar, and the custom distance performs better than both of them since it has learnable parameters. However, our distance metric consistently performs the best at all IoU thresholds. In our algorithm, we are using a Mahalanobis type distance function, and the metric in the distance function comes from the weights of the classification module. Although the custom metric has the capability, at least in theory, to learn the same metric as our proposed distance function, the direct coupling between the classification module and the metric learning module creates an extra boost in our algorithm that improves the performance.

\textbf{Effect of block-based processing.} We adopt a block-based processing strategy in the classification module to compute the classification score. In Table~\ref{tab:map_without_block}, we show the performance without block-based processing, i.e., when there is only one block for the whole video. From the experiment, we infer that block-based processing can handle variable length video more effectively. We still achieve superior performance compared to the current state-of-the-art without any block-based processing, mostly due to the metric learning module.

\hspace{0.4\textwidth}

\begin{table}[htbp]
    \centering
        \caption{The mAP performance at different IoU thresholds on the THUMOS14 dataset without any block-based processing in the classification module.}
    \begin{tabular}{|c |c c c c|}
    \hline
         IoU & 0.1 & 0.3 & 0.5 & 0.7 \\
         \hline
         mAP & 59.0 & 43.2 & 25.5 & 7.9 \\
         \hline
    \end{tabular}
    \label{tab:map_without_block}
\end{table}

\begin{figure}[htbp!]
    \begin{subfigure}[t!]{0.24\textwidth}  
            \includegraphics[width=\textwidth]{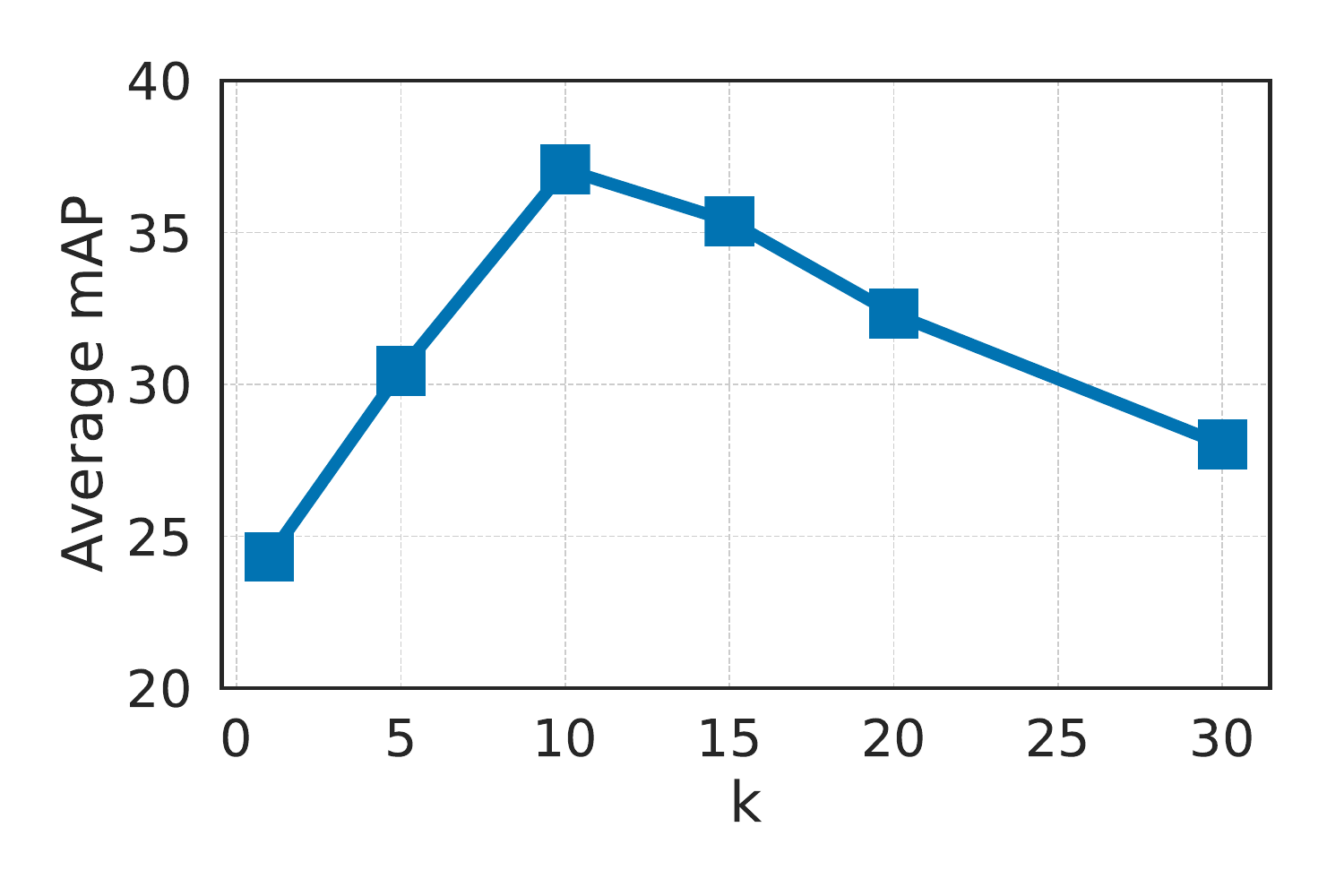}
            \vspace{-0.6cm}
            \caption{}
            \label{fig:k_value}
    \end{subfigure}%
    \begin{subfigure}[t!]{0.24\textwidth}
            \includegraphics[width=\textwidth]{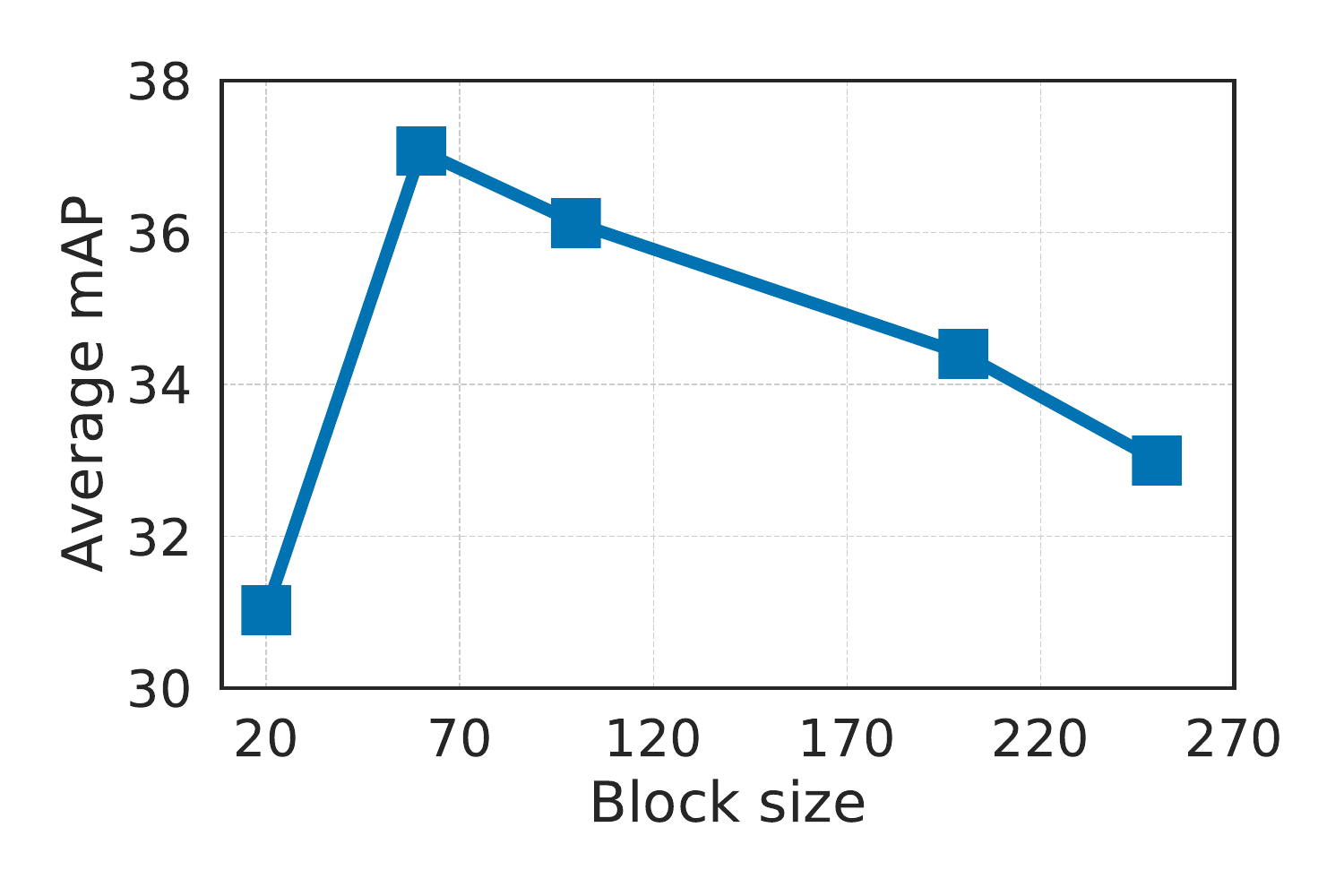}
            \vspace{-0.6cm}
            \caption{}
            \label{fig:block_size}
    \end{subfigure}
\vspace{-0.3cm}
    \caption{(a) The effect of $k$ for a fixed block size 60 on average mAP. (b) Variations of average mAP for different values of block size (here, $k$ is 6\% of the block size). The average mAP is calculated by averaging the mAPs for IoU thresholds 0.1, 0.3, 0.5, and 0.7.}
\end{figure}

\textbf{Effect of block size and $k$ value.} The block size $l_w$ and value of $k$ for $k$-max class activation are important parameters in our model (see Sec.~\ref{subsec:class}). The value of $k$ determines how many segments should be considered in each block to calculate the class score. From Fig.~\ref{fig:k_value}, we see that at $k=10$ for block size $60$, we get the highest average mAP. As $k$ increases or decreases, the performance degrades. The reason is that at lower $k$, noisy segments can corrupt the classification score, and at higher $k$, the model cannot detect very short-range action instances properly. Fig.~\ref{fig:block_size} illustrates the effect of block size $l_w$ on the final performance. Here, we again see that there is a trade-off for the value of $l_w$, and we get the best performance at around $l_w=60$.

\textbf{Ablation on clipping threshold.} Through experiments, we found that applying a clipping function $\varphi_\kappa(\cdot)$ increases the performance. In Table \ref{tab:abl_clip}, we show the mAP performance for different values of clipping thresholds $\kappa$, where `w/o clip' denotes the model where no clipping function $\varphi_\kappa(\cdot)$ is applied (or the threshold $\kappa$ is set to infinity). In particular, we obtain 2.5\% mAP improvement at IoU threshold 0.5 for $\kappa=4$ over no clipping. 

\begin{table}[!htb]
\caption{Experiments on clipping value $\kappa$}
\vspace{-0.3cm}
    \centering
    \begin{tabular}{|c|l l l l|}
    \hline

      \multirow{2}{*}{Clipping value $\kappa$} & \multicolumn{4}{c|}{IoU}\\ \cline{2-5}
    & 0.1 & 0.3 & 0.5 & 0.7   \\
    \hline
    w/o clip & 60.3 & 45.0 & 27.1 & 9.2 \\
    2 & 60.5 & 45.4 & 26.8 & 9.3 \\
    3 & 61.8 & 46.2 & 28.7 & 9.4 \\
    4 & \bf62.3 & \bf46.8 & \bf29.6 & \bf9.7 \\
    5 & 61.1 & 46.3 & 28.0 & 9.4 \\
    10 & 62.1 & 46.1 & 27.6 & 8.7 \\
    \hline
    \end{tabular}
    \label{tab:abl_clip}
    \vspace{-0.1cm}
\end{table}

\begin{figure*}[htbp!]
\begin{center}

    \begin{subfigure}[b]{\textwidth}            
            \includegraphics[width=\textwidth]{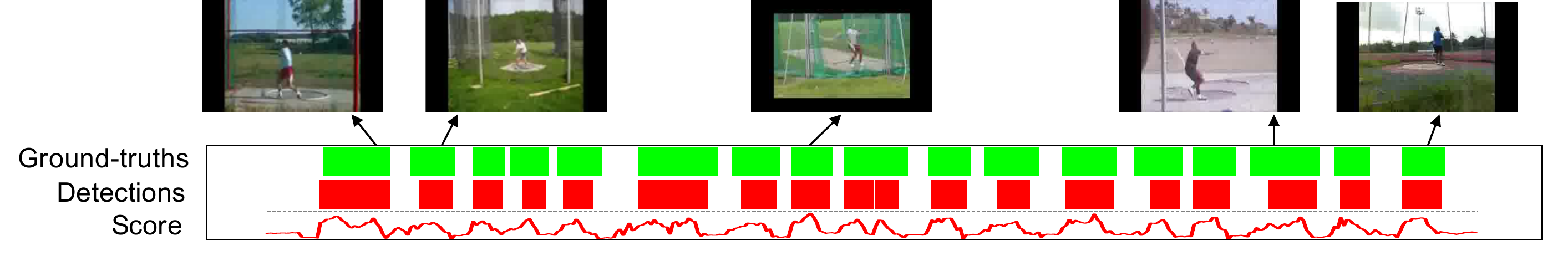}
            \vspace{-0.6cm}
            \caption{Hammer Throw}
            \label{fig:hammer}
    \end{subfigure}

    \begin{subfigure}[b]{\textwidth}
    \vspace{0.3cm}
            \includegraphics[width=\textwidth]{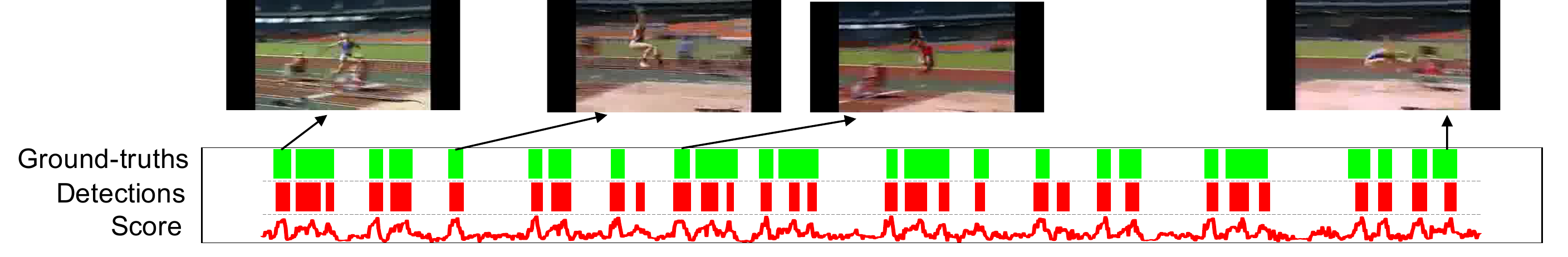}
            \vspace{-0.6cm}
            \caption{Long Jump}
            \label{fig:longjump}
    \end{subfigure}

        \begin{subfigure}[b]{\textwidth}   
        \vspace{0.3cm}
            \includegraphics[width=\textwidth]{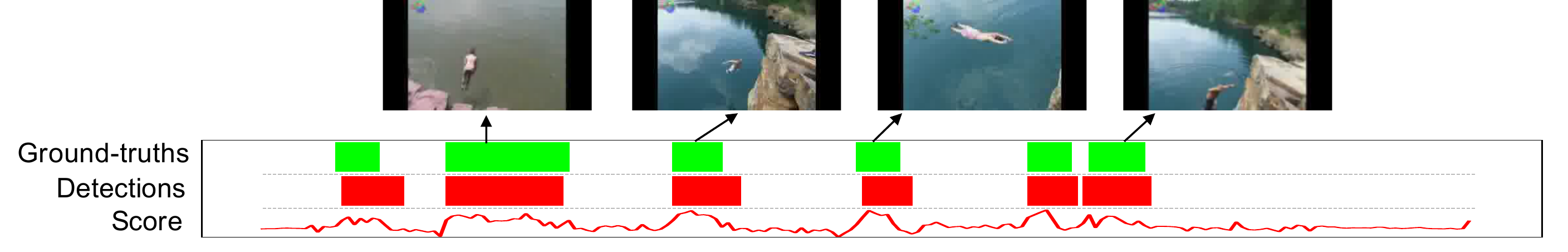}
            \vspace{-0.6cm}
            \caption{Cliff Diving}
            \label{fig:cliff}
    \end{subfigure}

    \begin{subfigure}[b]{\textwidth}
            \vspace{0.3cm}
            \includegraphics[width=\textwidth]{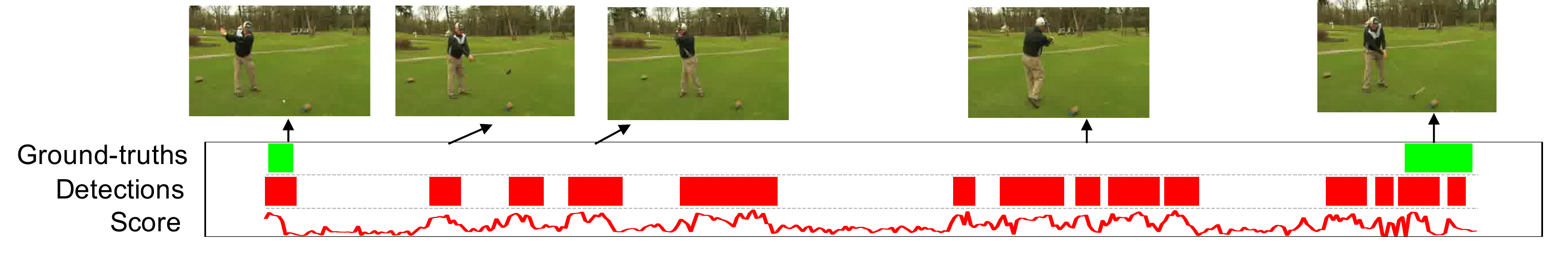}
            \vspace{-0.6cm}
            \caption{Golf Swing}
            \label{fig:golf}
    \end{subfigure}
\end{center}
\vspace{-0.5cm}
    \caption{Qualitative results on THUMOS14. The horizontal axis denotes time. On the vertical axis, we sequentially plot the ground truth detection, detection score after post-processing, and class activation score for a particular activity. (d) represents a failure case for our method. In (d), there are several false alarms where the person actually swings the golf club, but does not hit the ball.}
    \label{fig:qual}
\end{figure*}

\textbf{Qualitative results.} Figure~\ref{fig:qual} represents qualitative results on some videos from THUMOS14. In Fig.~\ref{fig:hammer}, there are many occurrences of the Hammer Throw activity, and due to the variation in background scene in the same video, it is quite challenging to localize all the actions. We see that our method still performs quite well in this scenario. In Fig.~\ref{fig:longjump}, the video contains several instances of the Long Jump activity. Our method can localize most of them effectively. Our method also localizes most activities in Fig.~\ref{fig:cliff} fairly well. Fig.~\ref{fig:golf} shows an example where our algorithm performs poorly. In Fig.~\ref{fig:golf}, there are several cases where the person swings the golf club or prepares to swing, but does not hit the ball. It is very challenging to differentiate actual Golf Swing and fake Golf Swing without any ground truth localization information. Despite several false alarms, our model still detects the relevant time-stamps in the video. 

\section{Conclusions and Future Work}\label{sec:conclusion}
We presented a weakly-supervised temporal action localization algorithm that predicts action boundaries in a video without any temporal annotation during training. Our approach achieves state-of-the-art results on THUMOS14, and competitive performance on ActivityNet1.2.  For action boundary prediction,  we currently rely on thresholding in the post-processing step. In the future, we would like to extend our work to incorporate the post-processing step directly into the end-to-end model.

\section{Acknowledgement}
This material is based upon work supported by the U.S.~Department of Homeland Security under Award Number 2013-ST-061-ED0001. The views and conclusions contained in this document are those of the authors and should not be interpreted as necessarily representing the official policies, either expressed or implied, of the U.S. Department of Homeland Security.

{\small
\bibliographystyle{ieee}
\bibliography{main_bibliography}
}
\end{document}